\documentclass[sigconf]{acmart}
\renewcommand\footnotetextcopyrightpermission[1]{}
\usepackage{float}
\usepackage{subfigure}
\usepackage{hyperref}
\usepackage{threeparttable}
\usepackage{color, colortbl}
\usepackage{multirow}
\usepackage{rotating}
\usepackage{arydshln}
\usepackage{makecell}

\newcommand{\tabincell}[2]{\begin{tabular}{@{}#1@{}}#2\end{tabular}}

\AtBeginDocument{%
  }

\begin{document}

\title{rLLM: Relational Table Learning with LLMs
}

\author{Weichen Li$^{1}$, Xiaotong Huang$^{1}$, Jianwu Zheng$^{1}$, Zheng Wang$^{1}$, Chaokun Wang$^{2}$, Li Pan$^{1}$, Jianhua Li$^{1}$
}
\affiliation{
    $^1$ Shanghai Jiao Tong University
    \country{China}
}
\affiliation{
    $^2$ Tsinghua University
    \country{China}
}
\email{{wzheng, panli, lijh888}@sjtu.edu.cn, chaokun@tsinghua.edu.cn}

\begin{abstract}
We introduce \textbf{rLLM} (relationLLM), a PyTorch library designed for Relational Table Learning (RTL) with Large Language Models (LLMs). 
The core idea is to decompose state-of-the-art Graph Neural Networks, LLMs, and Table Neural Networks into standardized modules, to enable the fast construction of novel RTL-type models in a simple "combine, align, and co-train" manner.
To illustrate the usage of rLLM, we introduce a simple RTL method named \textbf{BRIDGE}. Additionally, we present three novel relational tabular datasets (\textbf{TML1M}, \textbf{TLF2K}, and \textbf{TACM12K}) by enhancing classic datasets. 
We hope rLLM can serve as a useful and easy-to-use development framework for RTL-related tasks.
Our code is available at: https://github.com/rllm-project/rllm.
\end{abstract}



\begin{teaserfigure}
  \centering
  \begin{subfigure}{}
    \subfigure {
    \label{fig:a}     
    \includegraphics[width=0.48\textwidth]{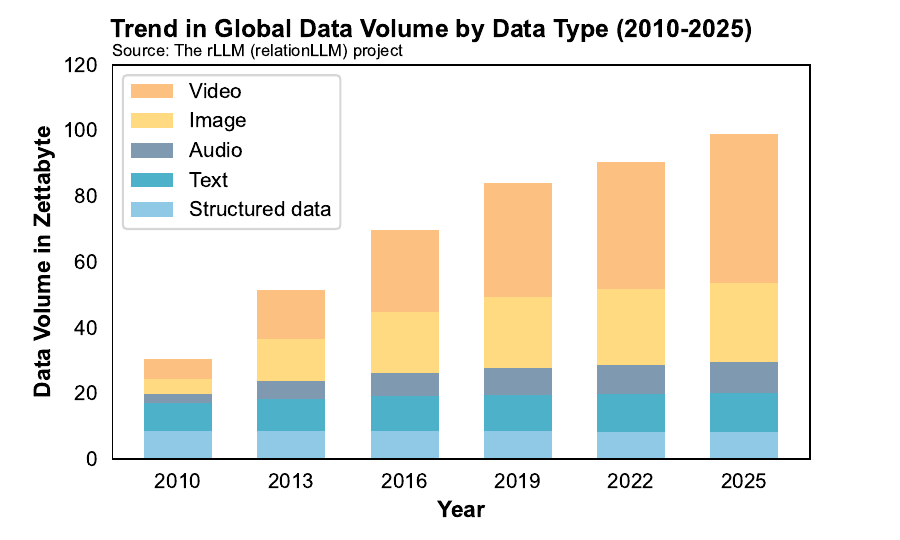}  
    }
    \subfigure{
    \label{fig:b}     
    \includegraphics[width=0.48\textwidth]{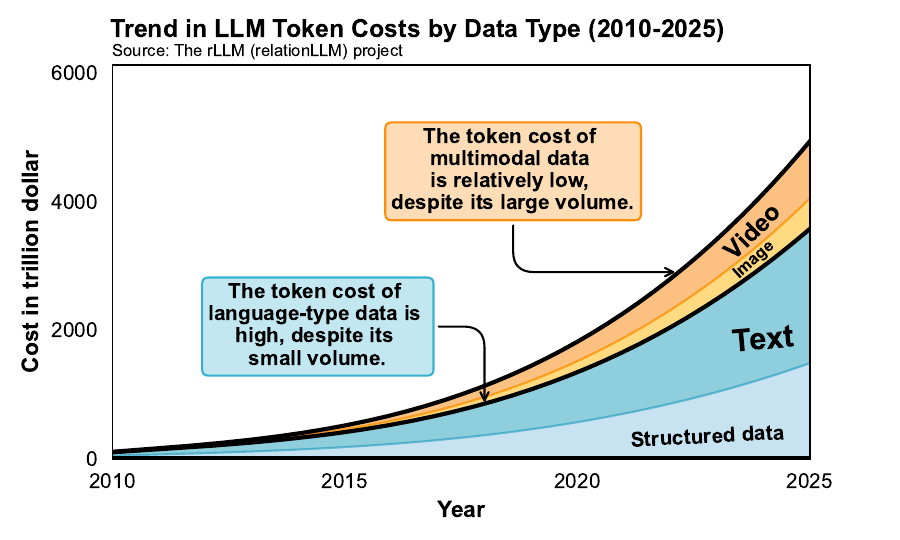}  
    }
  \end{subfigure}
  \caption{Trends in global data volume and in LLM token costs by data type}
  \label{trend}
\end{teaserfigure}

\maketitle

\section{Introduction}
Large language models (LLMs), like ChatGPT~\cite{brown2020language}, are driving a new wave of artificial intelligence advancements, garnering widespread attention. These models are great at 
at understanding and generating text by leveraging technologies such as web-scale unsupervised pretraining~\cite{zhao2023survey}, instruction fine-tuning~\cite{wei2022chain}, and value alignment~\cite{wolf2023fundamental}. 
Their strong performance across various tasks suggests they could be key to achieving Artificial General Intelligence~\cite{bubeck2023sparks}.

However, applying LLMs to real-world big data is extremely costly.
Figure 1 illustrates the growth trends of various types of data alongside the corresponding expenses\footnote{See Appendix 1 for details.}. The costs associated with LLMs are unsustainable.
For example, by 2025, the total cost of LLMs is anticipated to reach nearly \$5,000 trillion, which is 214 times the 2023 GDP of the United States (\$27.37 trillion).
Another notable observation is that processing text and structured data will incur most of these expenses, despite these two types of data being smaller in volume compared to multimedia data.

As relational databases host around 73\% of the world data~\cite{dbranking}, recent years have seen a significant shift towards Relational Table Learning (RTL)~\cite{dvzeroski2010relational}~\cite{zahradnik2023deep}~\cite{fey2023relational} \footnote{Although ``tabular'' is also commonly used to describe ``row-format data samples in relational databases'', our experience during development has indicated that undergraduate/graduate students, engineers, and business clients generally prefer the term ``table'' over ``tabular''.
This preference might be due to their more vocal nature. 
To facilitate public understanding, this paper uses the term ``table''.
}. 
In this paper, we introduce the \textbf{rLLM} (relationLLM) project, which aims to provide a platform for rapidly developing RTL-typle methods with LLMs. As shown in Figure~\ref{system_overview}, it performs two key functions: 1) decomposing state-of-the-art Graph Neural Networks (GNNs), LLMs, and Table Neural Networks (TNNs) into standardized modules, and 2) enabling the construction of novel models in a ``combine, align, and co-train'' manner using these decomposed modules.

To illustrate the application of rLLM, we present a simple illustrate RTL method named \textbf{BRIDGE}. 
The core idea is to streamline the workflow of RDL~\cite{fey2023relational} through two key simplifications: (1) considering only a single relational table connected to the target table, and (2) preprocessing all other tables into dense embeddings.
Specifically, BRIDGE employs TNNs to process the target table data and leverages the foreign-key links in the selected relational table to establish relationships among table samples, which are subsequently analyzed using GNNs.
This design efficiently captures inter-table dependencies with minimal architectural complexity.

Furthermore, as RTL is an emerging field with a notable lack of datasets, we further introduce a novel data collection (named \textbf{SJTUTables}) which includes three novel relational table datasets: \textbf{TML1M}, \textbf{TLF2K}, and \textbf{TACM12K}. 
Each dataset is obtained by enhancing existing classical datasets, and is accompanied by a standard classification task.
In other words, three datasets are well-organized and easy-to-use, making them very suitable for deigning novel RTL-type methods.
Future researchers are highly encouraged to use these datasets to define and develop additional tasks.

The structure of this paper is as follows: Section 2 details the framework of our rLLM system. Section 3 introduces the simple RTL algorithm BRIDGE, designed based on our project. Section 4 presents the methods currently supported by rLLM and describes the three new datasets. Section 5 provides a brief experimental comparison, and Section 6 concludes with a summary and directions for future research.

\section{System Overview}
\begin{figure}[t]
  \centering
  \includegraphics[width=0.9\linewidth]{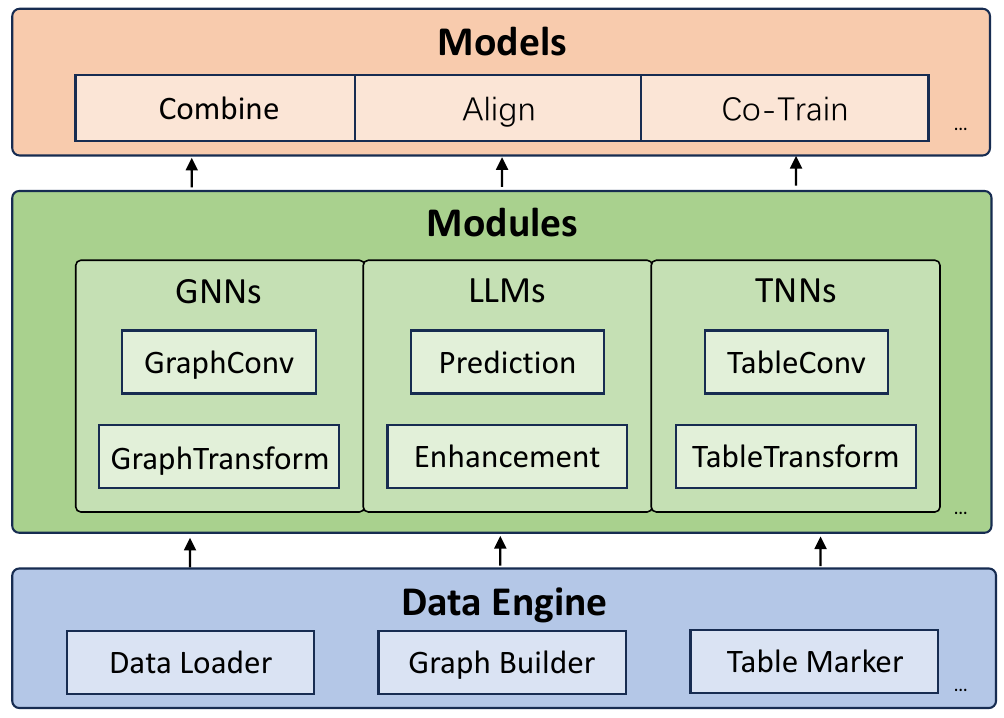}
  \caption{The architecture of rLLM}
  \label{system_overview}
\end{figure}

As shown in Figure~\ref{system_overview}, rLLM consists of three main layers: Data Engine Layer, Module Layer, and Model Layer.

\subsection{Data Engine Layer}
\begin{figure}[!t]
  \centering
\includegraphics[width=\linewidth]{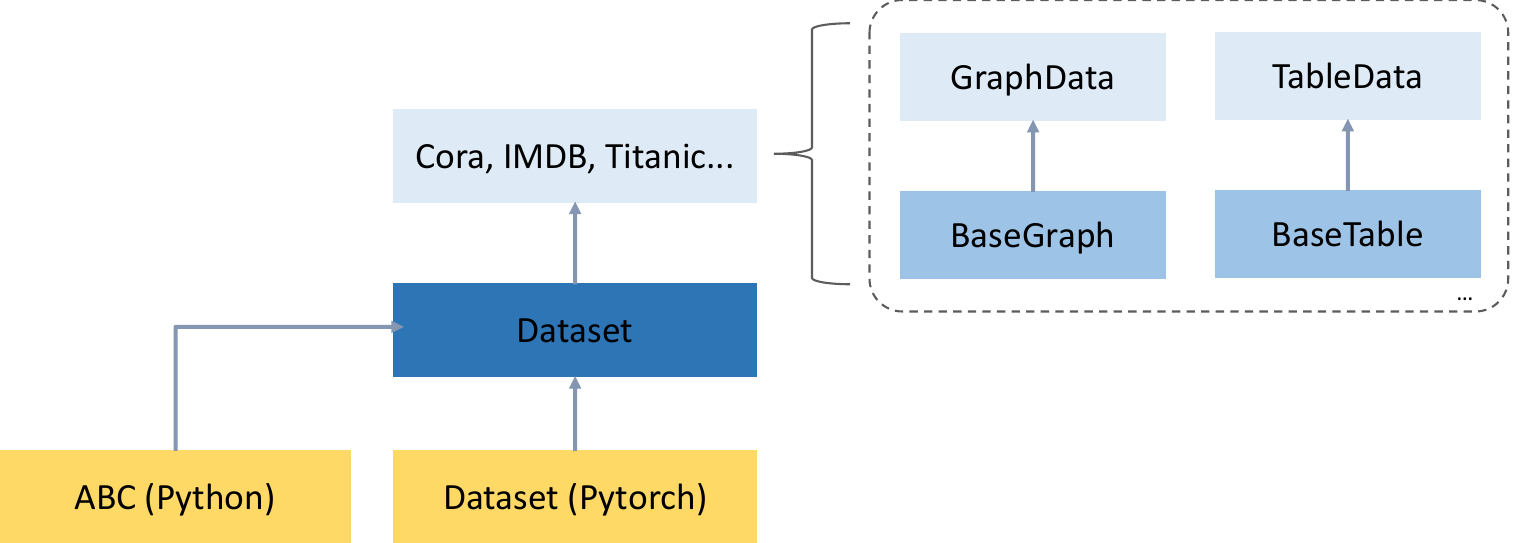}
  \caption{Base data structure in rLLM. Arrows indicate inheritance relationships and parentheses indicate containment relationships.}
  \label{data_engine}
\end{figure}

This layer designs the fundamental data structures for graph and table data and defining the processing workflows for relational table data. 
As shown in Figure~\ref{data_engine}, the overall architecture decouples data loading and storage, which are handled by the \textit{Dataset} subclass and \textit{BaseGraph}/\textit{BaseTable} subclasses, respectively. This design philosophy is driven by the pursuit of flexibility and scalability, enabling efficient and flexible handling and storage of different graph and table data.

Specifically, within the Dataset subclasses, we can implement various data loading classes tailored to the characteristics of different datasets. These subclasses focus on efficiently data loading and preprocessing, ultimately storing the processed data in data structures inherited from \textit{BaseGraph} and \textit{BaseTable}. Additionally, they are both optimized for the storage and processing of graph data (such as homogeneous and heterogeneous graphs) and table data, respectively. Overall, this design meets the familiar storage and processing requirements of relational table data consist of table data and foreign key relationships.

\subsection{Module Layer}
This layer decomposes the operations of GNNs, LLMs, and TNNs into standard submodules.

\subsubsection{GNN Modules}
\label{sub_sect_gnn_module}
This part mainly includes the \textit{GraphTransform} and \textit{GraphConv} modules. The \textit{GraphTransform} module provides preprocessing methods for graph data, such as normalization and self-loop operations. Additionally, this module supports combining various graph preprocessing methods, allowing users to perform complex graph preprocessing operations to meet the requirements of subsequent algorithms.

The \textit{GraphConv} module implements popular graph convolution layers, including homogeneous and heterogeneous graph convolutions. 
The core functionality of this module involves different message-passing functions between nodes for various graph convolution operations, addressing the requirements of diverse graph types and algorithms. In practical applications, stacking and interacting multiple graph convolution layers allows for the modeling of complex graph data information.

\begin{figure*}
    \centering
    \includegraphics[width=0.7\textwidth]{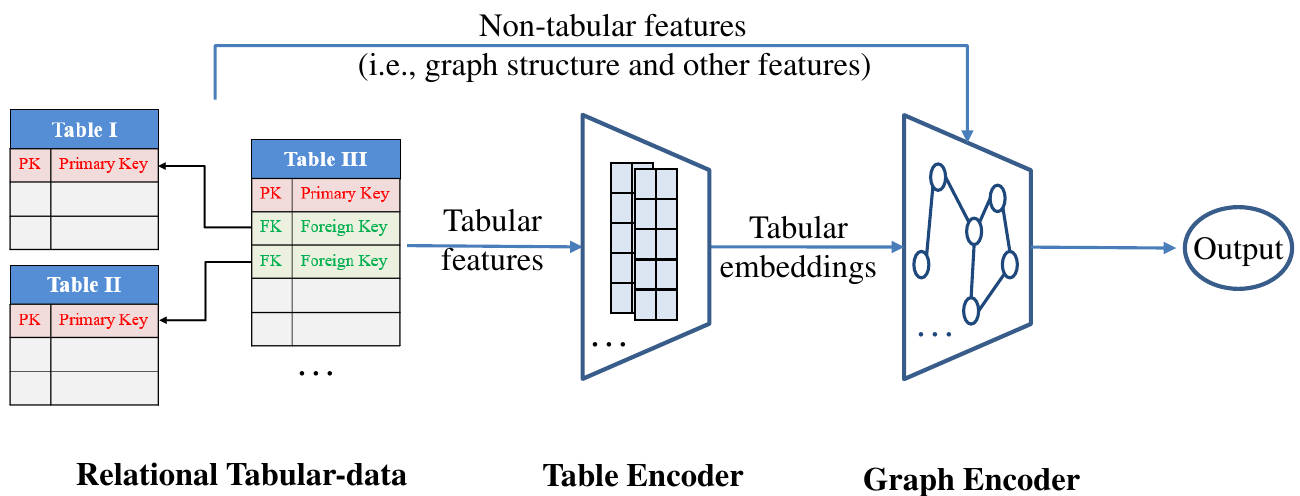}
    \caption{The architecture of BRIDGE}
    \label{fig:bridge}
\end{figure*}

\subsubsection{LLM Modules}
\label{sub_sect_llm_module}
This part mainly includes the \textit{Predictor} and \textit{Enhancer} modules. The \textit{Predictor} module allows users to leverage LLMs for data annotation. It is crucial because much real-world data lacks labels, and manual annotation is expensive and prone to errors. Users can use existing prompts or design their own to enable LLMs to perform preliminary annotations. These annotations can serve as final label predictions or as labeled data for further training of other machine learning models.

The \textit{Enhancer} module allows users to employ LLMs for data augmentation. In many real-world scenarios, data may be insufficient or of low quality. For example, understanding the full context of a research paper based solely on its title can be challenging. Users can use various prompts to generate detailed textual explanations for data samples with the LLM. These explanations can be used directly to enhance the data or be transformed into other feature formats to improve performance in downstream tasks.

\subsubsection{TNN Modules}
\label{sub_sect_tnn_module}
This part mainly includes the \textit{TableTransform} and \textit{TableConv} modules. The \textit{TableTransform} module maps sample features to higher-dimensional vector spaces. It is necessary because table data samples (i.e., rows) consist of multiple feature columns, which can vary greatly in nature. Due to the diverse types of features and the often limited information provided by tables with fewer columns, it is crucial to map or transform some columns into higher-dimensional feature spaces to enhance the sample information.
The \textit{TableConv} module facilitates multi-layer interactive learning among feature columns to extract latent information. Since the contribution of each column to downstream tasks varies (e.g., blood sugar levels in a diabetes dataset significantly impact diabetes prediction), this module usually employs various attention mechanisms to automatically learn and extract complex relationships from features. In reality, stacking and interacting multiple layers of this module  generally provide sufficient capacity for comprehensive learning from table data.

\subsection{Model Layer}
By combining modules from the second layer, this layer provides three main strategies for rapidly developing RTL-type models: \textbf{Combine}, \textbf{Align}, and \textbf{Co-Train}.

\begin{itemize}
    \item \textbf{Combine} refers to jointly using modules from different parts. For example,~\cite{perozzi2024let} and~\cite{chen2023label} use LLMs for preliminary label annotation, and then the annotated results are fed into a GCN~\cite{gcn} model for graph node classification. Within the rLLM framework, we can utilize the Predictor module from the LLM part to complete the annotation task and then use the GCN module from the GNN part for the following classification task.
    \item \textbf{Align} involves aligning the input and output feature spaces for different modules. For instance, ConGraT~\cite{brannon2023congrat} generates node embeddings using a language model and a GNN separately, followed by aligning these embeddings within the final embedding space. Within the rLLM framework, we can first use the Enhancer module from the LLM part to generate embeddings (and potentially perform enhancement simultaneously), then generate node embeddings with the GNN module, and finally align the two types of embeddings.
    \item \textbf{Co-Train} denotes the collaborative training of different modules. For example, BRIDGE (Section~\ref{sec:bridge})  integrates TNNs and GNNs to leverage internal and inter-table information. Within the rLLM, we can invoke  modules from both GNNs and TNNs, combine them as needed, and perform co-training to enhance the performance of multi-table joint learning tasks.
\end{itemize}

Moreover, these three strategies can be employed independently or in combination, allowing for the rapid development of various RTL-type models.
\section{An Illustration Method - BRIDGE}
\label{sec:bridge}
In this section, we introduce a straightforward method named the \textbf{B}asic \textbf{R}elat\textbf{I}onal table-\textbf{D}ata Learnin\textbf{G} Fram\textbf{E}work (\textbf{BRIDGE}) to illustrate how to quickly construct RTL-type methods using rLLM. 
The core idea of BRIDGE is to simplify the workflow of RDL~\cite{fey2023relational} through two key simplifications: (1) only a single relational table connected to the target table is considered, and (2) all other tables, except for the target and selected relational table, are preprocessed into dense embeddings.
The details are as follows.

Specifically, in real-world applications of relational databases, data is usually stored in multiple (two-dimensional) tables connected by foreign keys. This means relational table data includes table features (i.e., each sample is a row in a table with many columns) and non-table features (such as the graph structure formed by foreign key relationships). Therefore, we need to handle both the table data and their interrelationships.

Firstly, for the target table data, we need to use table neural networks to model and learn from the data. This is due to the heterogeneity of table data, where each column's feature type and meaning can vary significantly. As shown in Figure \ref{fig:bridge}, we use a Table Encoder to model the table features. As mentioned in Section \ref{sub_sect_tnn_module}, we can construct different Table Encoders using the \textit{TableTransform} and \textit{TableConv} modules from the TNN part in rLLM, to obtain the table embeddings.

Secondly, for the selected relational table, which encodes foreign-key relationships, we construct associations among samples and model them with a Graph Encoder.
As described in Section \ref{sub_sect_gnn_module}, we can use the \textit{GraphTransform} and \textit{GraphConv} modules from the GNN part in rLLM, to construct various Graph Encoders, ultimately producing graph embeddings. 
Additionally, target table embeddings from the TNN, along with preprocessed embeddings of other tables, are also input to the Graph Encoder for joint modeling.

Finally, we integrate the outcomes from the table encoder and the graph encoder, allowing the BRIDGE framework to concurrently model multi-table data and their interconnections. The training objective of the entire model can be either supervised (e.g., using a cross-entropy loss function based on labeled data) or unsupervised (e.g., designing an unsupervised loss function based on data reconstruction principles from table data or foreign key relationships). 
\begin{table*}[!t]
\caption{Summary of the datasets.}
\small
\label{tab_datasets}
\begin{center}
    \begin{tabular}{l c c c c c}
    \toprule
    \textbf{Dataset} & \textbf{Tables [\#row/\#col]} & \textbf{Relation\ Tables} & \textbf{Label} &  \textbf{Classes} & \textbf{\#Train/\#Val/\#Test} \\
    \midrule
    {TML1M}  & \tabincell{c}{users [6,040/5]\\movies [3,883/11]\\ratings [1,000,209/4]} & \tabincell{c}{ratings: user-movie} & Age range of user & 7 & [140/500/1000]  \\
    \midrule
    {TLF2K}  & \tabincell{c}{artists [9,047/10]\\user\_artists [80,009/3]\\user\_friends [12,717/3]} & \tabincell{c}{user\_artists: user-artist\\user\_friends: user-user} & Genre of artist & 11 & [220/500/1000] \\
    \midrule
    {TACM12K} & \tabincell{c}{papers [12,499/5]\\authors [17,431/3]\\citations [30,789/2]\\writings [37,055/2]} & \tabincell{c}{citations: paper-paper\\writings: paper-author} & Conference of paper
 & 14 &[280/500/1000] \\
    \bottomrule
    \end{tabular}
    \label{tab_sod}
\end{center}
\end{table*}

\section{Methods and Datasets}
\subsection{Included methods.}
We have already included some common methods, which can be categorized into two types.
\begin{enumerate}
    \item GNN-type Methods:
    \begin{itemize}
        \item Homogeneous Methods: GCN~\cite{gcn}, GAT~\cite{gat}, RECT~\cite{rect}, TAPE~\cite{tape}, and OGC~\cite{wang2023cluster}.
        \item HeterogeneousMethods: HAN~\cite{han} and HGT~\cite{hu2020heterogeneous}.
    \end{itemize}
    \item TNN-type Methods:
    \begin{itemize}
        \item Single-Table Learning: TabTransformer~\cite{tabtransformer}, TabNet~\cite{tabnet}, and FT-Transformer~\cite{ft-transformer}.
        \item Relational Table Learning: BRIDGE.
    \end{itemize}
\end{enumerate}

\subsection{Included Datasets}
In addition to the common graph data and single-table data, we further introduce a novel data collection (named \textbf{SJTUTables}) which includes three novel relational table datasets by
enhancing existing classical ones. 
As summarized in Table~\ref{tab_datasets}, each dataset has a default classification task with a balanced and fixed train/val/test split setting.
Specifically, we provide 20 labeled samples for each class in these datasets, with two additional sets of 500 validation samples and 1000 test samples.
The complete details are as follows.
\begin{itemize}
    \item \textbf{Table-MovieLens1M (TML1M)} is a relational table dataset enhanced from the classical MovieLens1M dataset\footnote{https://grouplens.org/datasets/movielens/1m/}, comprising three tables: \textit{users}, \textit{movies} and \textit{ratings}. We have enriched the movie table with more comprehensive features and defined a new task for classifying user age ranges.
    \item \textbf{Table-LastFm2K (TLF2K)} is a relational table dataset enhanced from the classical LastFm2k dataset\footnote{https://grouplens.org/datasets/hetrec-2011/}, containing three tables: \textit{artists}, \textit{user\_artists} and \textit{user\_friends}. We have enriched the artists table with more detailed features and streamlined the tags for each artist, defining a new task for music genre classification of artists.
    \item \textbf{Table-ACM12K (TACM12K)} is a relational table dataset enhanced from the ACM heterogeneous graph dataset\footnote{https://github.com/Jhy1993/HAN/tree/master/data/acm}. It includes four tables: \textit{papers}, \textit{authors}, \textit{citations} and \textit{writings}. The paper features include year, title, and abstract, while author features include name and affiliation. 
    In addition, we also make some completion when some features of papers are lost.
    The task is to predict the conference of papers.
\end{itemize}
\textbf{Compared to other datasets.} The RelBench dataset~\cite{fey2023relational}, recently scraped from websites like Amazon and Stack Overflow, includes commercially practical applications such as user lifetime value prediction and user forum activity prediction. However, this practicality also introduces some additional challenges such as imbalance, high noise levels, complex tasks, and large data volumes. 
In contrast, SJTUTables are enhancements of classic datasets, offering simplicity and reliable data quality. In addition, its datasets focus on the most classic classification tasks in the machine learning field, providing fixed, balanced splits to facilitate standardized evaluation. Therefore, we strongly recommend designing standard RTL methods based on SJTUTables before conducting further commercial application evaluations on the RelBench dataset.
\section{Evaluation}
\subsection{Experimental Setup}
To demonstrate our system and the proposed BRIDGE algorithm, we conducted comparative experiments on the TML1M dataset. In the BRIDGE algorithm, we used TabTransformer as the table encoder and GCN as the graph encoder. To ensure fairness, we standardized training batches, dropout rates, and other parameters for each method and conducted multiple experiments to get the average results.
\subsection{Comparative Methods}
We utilized the following baselines built into rLLM\footnote{Due to the inability of the GNNs to directly process tabular data, GNN-type methods are not compared in this experiment.}:
\begin{itemize}
  \item TabTransformer: A deep tabular data modeling architecture for supervised and semi-supervised learning. This model is based on a self-attention Transformer architecture.    
  \item  TabNet: A model for Attentive Interpretable Tabular Learning, which uses neural networks to build a structure akin to decision trees for analyzing tabular data.
  \item FT-Transformer: A simple Transformer adaptation for the tabular data domain. The model's Feature Tokenizer component converts all features (categorical and numerical) into tokens and adds a [CLS] token. 
\end{itemize}

\subsection{Results and Analysis}

\begin{table}[!t]
\small
\caption{Classification accuracy.}
\centering
\begin{tabular}{l|ccc}
\toprule
Methods$\backslash$Datasets & TML1M & TLF2K & TACM12K  \\
\hline
Random & 0.144$\pm$0.01 & 0.091$\pm$0.03 & 0.075$\pm$0.0\\
TabTransformer & 0.347$\pm$0.02 & 0.1370$\pm$0.08 & 0.091$\pm$0.01\\
TabNet & 0.259$\pm$0.08 & 0.1346$\pm$0.03 & 0.135$\pm$0.01\\
FT-Transformer & 0.352$\pm$0.02 & 0.1319$\pm$0.01 & 0.099$\pm$0.01\\
BRIDGE 	&\textbf{0.362$\pm$0.03}	&\textbf{0.422$\pm$0.03}	&\textbf{0.256$\pm$0.01}\\
\bottomrule
\end{tabular}
\label{tab:eva_result}
\end{table}

The experimental results indicate that traditional single-tabular TNNs can only learn knowledge from the single target table, failing to effectively utilize the information provided by multiple tables and the relationships between them. Consequently, their performance is relatively poor. In contrast, the BRIDGE algorithm effectively extracts valuable information from both the various tables and the relationships between them by combining a table encoder and a graph encoder, leading to a notable improvement in performance.

\section{Conclusion}
We presented rLLM framework for relational table learning with LLMs. We are actively working to further integrate more advanced methods and also plan to optimize relevant data structures to improve system efficiency. All researchers and software engineers are welcomed to collaborate with us in extending its scope.

\bibliographystyle{unsrt}
\bibliography{main_rllm}

\appendix
\section{Appendix}
\subsection{Details of Figure~\ref{trend}}

\subsubsection*{\textbf{Global Data Trend Estimation}}
We calculated the total data distribution based on statistics and projections from Statista~\cite{statista2021worldwide}, which estimate that the global data volume will reach a staggering 181ZB by 2025. Additionally, using reports from Chopra ~\cite{chopra2015big} and IBM~\cite{guide2013getting}, we roughly estimated the proportion of each modality and calculated their corresponding data volumes based on the projected global data total.
\subsubsection*{\textbf{LLM Cost Trend Estimation}}
To calculate the LLM cost, we first determined the number of tokens that different modalities of data could be converted into. First, for pure text data, we assumed the text to be in UTF-8 encoded plain English, where each character occupies one byte. According to relevant studies~\cite{smith2012distinct}, English words are approximately 6-11 characters. Using this ratio, we calculated the number of words corresponding to the text data and then used OpenAI's empirical formula~\cite{open2023tokens} to estimate the number of tokens. Second, for multimodal data, such as images and videos, we applied different methodologies. For images, we used 256px by 256px RGB images and employed tokenization techniques similar to those used in Vision Transformer~\cite{dosovitskiy2020image}, in conjunction with OpenAI's pricing strategy for estimation\cite{open2024pricing}. For videos, we referred to Google's Gemini 1.5 technical report~\cite{reid2024gemini}, analyzing token calculation strategies demonstrated in the AlphaGo documentary. This approach involved randomly sampling one frame per second from the video and calculating the number of tokens in the same way as for images. Last, after determining the number of tokens for each modality, we calculated the associated costs. We assume that these data are uniformly given to GPT-3.5 Turbo[4] for analysis, the overhead required to output the token is ignored, and finally, the approximate analysis cost is obtained.

\subsection{Datasets}
We introduce a novel data collection (named \textbf{SJTUTables}) which includes three novel relational table datasets by enhancing existing classical ones.
Each dataset has a default single-label classification task with a balanced and fixed train/val/test split setting.
Specifically, we provide 20 labeled samples for each class in these datasets, with two additional sets of 500 validation samples and 1000 test samples.
The complete details are as follows.

\subsubsection{\textbf{TML1M}}
Derived from the classical MovieLens 1M dataset\footnote{https://grouplens.org/datasets/movielens/1m/}, the TML1M dataset consists of three relational tables.
\begin{itemize}
    \item \textbf{\textit{users} table}: Inherited from the MovieLens 1M dataset, it includes UserID, Gender, Age, Occupation and Zip-code information of users.
    In sum, the row number is 6,040 and the column number is 5.
    \item \textbf{\textit{movies} table}: Enhanced by scraping the MovieLens website to gather additional metadata for each movie, including MovieID, Title, Year, Genre, Director, Cast, Runtime, Languages, Certificate, Plot and Url. This enhancement provides more comprehensive and detailed textual information of movies.
    In sum, the row number is 3,883 and the column number is 11. 
    \item \textbf{\textit{ratings} table}: Contains UserID, MovieID, Rating, and Timestamp of ratings.
    In sum, the row number is 1,000,209 and the column number is 4.
\end{itemize}
The default task for this dataset is to predict the user's age range in the User table.
\subsubsection{\textbf{TLF2K}}
Derived from the classical LastFM 2K dataset\footnote{https://grouplens.org/datasets/hetrec-2011/}, the TLF2K dataset consists of three relational tables.
\begin{itemize}
    \item \textbf{\textit{artists} table}: Enhanced by scraping Last.FM, its column includes artistID, type, name, born, yearsActive, location, genre, tag\_list, biography and url.
    Every artist is classified into one of 11 genres.
    Specifically, ChatGPT was used to assign a single label to each artist based on their tag list, including confidence scores, with manual intervention for low-confidence labels. 
    As such, original tags, by default, should not be used for this classification task.
    In sum, the row number is 9,047 and the column number is 10.
    \item \textbf{\textit{user\_friends} table}: Contains 12,717 bi-directional edge relationships, representing friendships among 1,892 users.
    In sum, the row number is 12,717 and the column number is 2. The columns are userID an friendID.
    \item \textbf{\textit{user\_ artists} table}: Represents listening relationships between users and artists.
    In sum, the row number is 80,009 and the column number is 3. The columns are userID, artistID and weight, where weight represents listening count.
\end{itemize}
The default task for this dataset is to predict the music genre of artists.

\subsubsection{\textbf{TACM12K}}
Derived from the classical ACM heterogeneous graph dataset\footnote{https://github.com/Jhy1993/HAN/tree/master/data/acm}, TACM12K contains four tables:
\begin{itemize}
    \item \textbf{\textit{papers} table}: Manually labeled the year information for venues and added year attributes via the 'PvsV' matrix. Corrected errors in the original 'PvsC' markings, reclassified STOC as COLT in 'VvsC', and recalculated the 'PvsV * VvsC' matrix to add conference attributes to papers. 
    In addition, we also provide the title and abstract information as two columns. 
    In sum, the row number is 12,499 and the column number is 5, and the columns are paper\_id, year, conference, title and abstract.
    \item \textbf{\textit{authors} table}: Extracted from the file, it includes original author IDs and names, with firm information added based on 'AvsF'.
    In sum, the row number is 17,431 and the column number is 3, and the columns are author\_id, name and firm.
    \item \textbf{\textit{citations} table}: Derived from the original 'PvsP' matrix. 
    In sum, the row number is 30,789 and the column number is 2, and the columns are paper\_id and paper\_id\_cited, where the former column cites the latter.
    \item \textbf{\textit{writings} table}: Derived from the original 'PvsA' matrix. 
    In sum, the row number is 37,055 and the column number is 2, and the columns are paper\_id and author\_id.
\end{itemize}
The default task for this dataset is to predict the conference of papers.

\end{document}